\documentclass[journal]{IEEEtran}
\usepackage{cite}

\usepackage[font=small]{caption}
\usepackage{bm}
\usepackage{amssymb}
\usepackage{enumerate}
\usepackage{amsmath}
\usepackage{booktabs}
\usepackage{multirow}
\usepackage{subcaption}
\usepackage{lipsum}
\usepackage{hyperref}

\DeclareMathOperator*{\argmin}{arg\,min}
\ifCLASSINFOpdf
   \usepackage[pdftex]{graphicx}
\else
\fi
\hypersetup{
 linkcolor=red,
 citecolor=green,
 colorlinks=true
}

\begin{document}
%
\title{How would surround vehicles move?\\ A Unified Framework for Maneuver Classification and Motion Prediction }
%
%
%

\author{Nachiket~Deo,~\IEEEmembership{}
        Akshay~Rangesh,~\IEEEmembership{}
        and~Mohan~M.~Trivedi,~\IEEEmembership{Fellow,~IEEE}
\thanks{The authors are with the Laboratory for Intelligent and Safe Automobiles
(LISA), University of California at San Diego, La Jolla, CA 92093 USA}
\thanks{(email:ndeo@ucsd.edu, arangesh@ucsd.edu, mtrivedi@ucsd.edu)}
\thanks{Manuscript info.}}

%
%

\markboth{Journal name}%
{Shell \MakeLowercase{\textit{et al.}}: Bare Demo of IEEEtran.cls for IEEE Journals}
%



\maketitle
\global\csname @topnum\endcsname 0
\global\csname @botnum\endcsname 0
\begin{abstract}
Reliable prediction of surround vehicle motion is a critical requirement for path planning for autonomous vehicles. In this paper we propose a unified framework for surround vehicle maneuver classification and motion prediction that exploits multiple cues, namely, the estimated motion of vehicles, an understanding of typical motion patterns of freeway traffic and inter-vehicle interaction. We report our results in terms of maneuver classification accuracy and mean and median absolute error of predicted trajectories against the ground truth for real traffic data collected using vehicle mounted sensors on freeways. An ablative analysis is performed to analyze the relative importance of each cue for trajectory prediction. Additionally, an analysis of execution time for the components of the framework is presented. Finally, we present multiple case studies analyzing the outputs of our model for complex traffic scenarios.   
\end{abstract}


%
\IEEEpeerreviewmaketitle

\section{Introduction}

\begin{figure}[]
\centering
\includegraphics[width=2.9in]{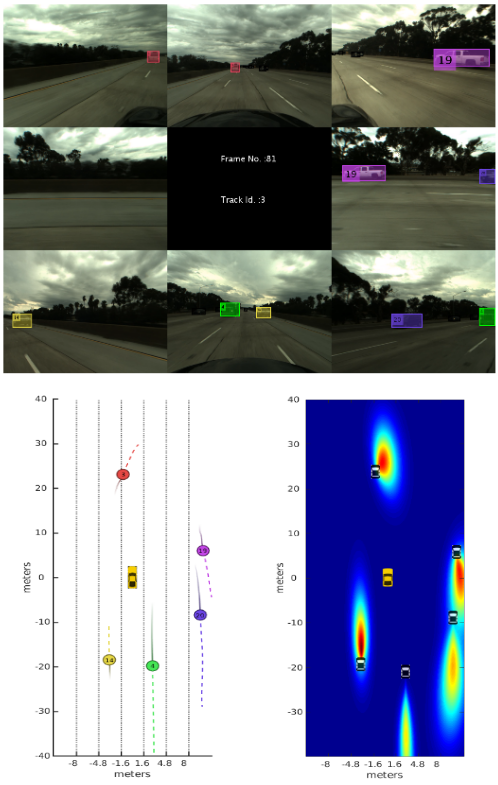}
\caption{ To smoothly navigate through freeways an autonomous vehicle must estimate a distribution over the future motion of the surrounding vehicles. We propose a unified model for trajectory prediction that leverages the instantaneous motion of the vehicles, the maneuver being performed by the vehicles and inter-vehicle interactions, while working purely with data captured using vehicle mounted sensors. The above figure shows the data captured by 8 surround cameras \textbf{(top)}, the track histories of surround vehicles, the mean predicted trajectories \textbf{(bottom left)} and a heat map of the predicted distribution in the ground plane \textbf{(bottom right)}.
}
\label{fig_pull}
\end{figure}
For successful deployment in challenging traffic scenarios, autonomous vehicles need to ensure the safety of its passengers and other occupants of the road, while navigating smoothly without disrupting traffic or causing discomfort to its passengers. Existing tactical path planning algorithms \cite{ulbrich,nilsson,dpdm} hinge upon reliable estimation of future motion of surrounding vehicles over a prediction horizon of up to 10 s. While approaches leveraging vehicle-to-vehicle communication \cite{v2v2,v2v3,v2v4}, offer a possible solution, these would require widespread adoption of autonomous driving technology in order to become viable. In order to safely share the road with human drivers, an autonomous vehicle needs to have the ability to predict the future motion of surrounding vehicles purely based on perception. Thus, we address the problem of surround vehicle motion prediction purely based on data captured using vehicle mounted sensors.

Prediction of surround vehicle motion is an extremely challenging problem due to a large number of factors that affect the future trajectories of vehicles. Prior works addressing the problem seem to incorporate three cues in particular: the instantaneous estimated motion of surround vehicles, an understanding of typical motion patterns of traffic and inter-vehicle interaction. A large body of work uses the estimated state of motion of surround vehicles along with a kinematic model to make predictions of their future trajectories \cite{motion1,motion2,motion3,motion4,motion5,motion6,imm1,imm2}. While these approaches are computationally efficient, they become less reliable for long term prediction, since they fail to model drivers as decision making entities capable of changing the motion of vehicles over long intervals. An alternative is offered by probabilistic trajectory prediction approaches \cite{vgmm,gp1,gpclass1,gpclass2,gmrlat,gmrlong} that learn typical motion patterns of traffic from a trajectory dataset. However these approaches are prone to poorly modeling safety critical motion patterns that are under represented in the training data. Many works address these shortcomings of motion models and probabilistic models by defining a set of semantically interpretable \textit{maneuvers} \cite{man1,man2,man3,man4,jacobmiklas,aida,houenou,schreier,anup}. A separate motion model or probabilistic model can then be defined for each maneuver for making future predictions. Finally some works leverage inter-vehicle interaction for making trajectory predictions \cite{vi1,vi2}.

While many promising solutions have been proposed, they seem to have the following limitations. (i) Most works consider a restrictive setting such as only predicting longitudinal motion, a small subset of motion patterns, or specific cases of inter-vehicle interaction, whereas many of the biggest challenges for vehicle trajectory prediction originate from the generalized setting of simultaneous prediction of the complete motion of all vehicles in the scene. (ii) Many approaches have been evaluated using simulated data, or based on differential GPS, IMU readings of target vehicles, whereas evaluation using real traffic data captured using perceptual vehicle mounted sensors is more faithful to the setting being considered. (iii) There is a lack of a unifying approach that combines each of the three cues mentioned above and analyzes their relative importance for trajectory prediction. 

In this work, we propose a framework for holistic surround vehicle trajectory prediction based on three interacting modules:  A hidden Markov model (HMM) based maneuver recognition module for assigning confidence values for maneuvers being performed by surround vehicles, a trajectory prediction module based on the amalgamation of an interacting multiple model (IMM) based motion model and maneuver specific variational Gaussian mixture models (VGMMs), and a vehicle interaction module that considers the global context of surround vehicles and assigns final predictions by minimizing an energy function based on outputs of the other two modules. We work with vehicle tracks obtained using 8 vehicle mounted cameras capturing the full surround and generate the mean predicted trajectories and prediction uncertainties for all vehicles in the scene as shown in Figure \ref{fig_pull}. We evaluate the model using real data captured on Californian freeways.

The main contributions of this work are:
\begin{enumerate}
\item A unified framework for surround vehicle trajectory prediction that exploits instantaneous vehicle motion, an understanding of typical motion patterns of traffic and inter-vehicle interaction.
\item An ablative analysis for determining the relative importance of each cue in trajectory prediction.
\item Evaluation based on real traffic data captured using vehicle mounted sensors.
\end{enumerate}

\section{Related Research}
\begin{figure*}[t!]
\captionsetup{}
\centering
\includegraphics[width=7in]{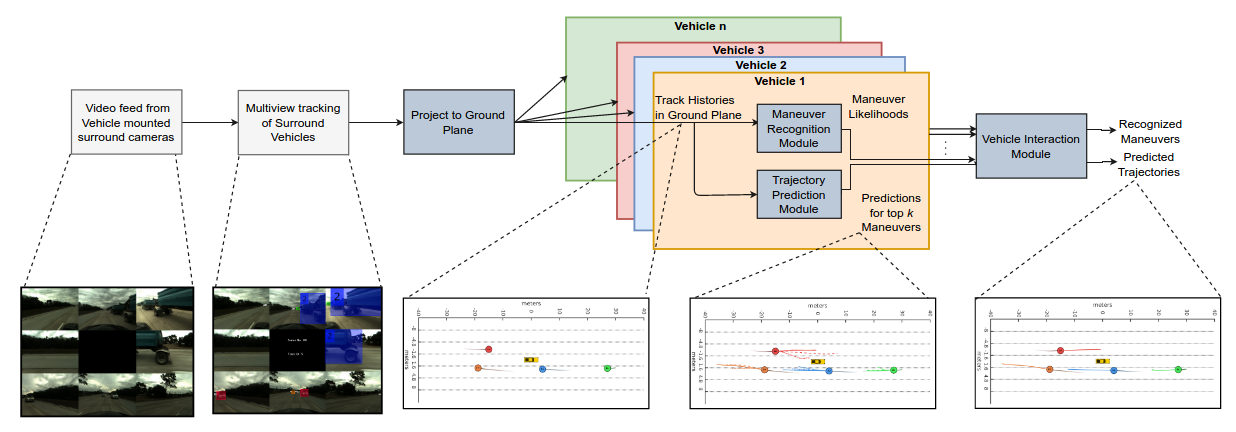}
\caption{\textbf{Overview of the proposed model:} Track histories of all surround vehicles are obtained via a multi-perspective tracker and projected to the ground plane in the ego vehicle's frame of reference. The model consists of three interacting modules: The maneuver recognition module assigns confidence values to possible maneuvers being performed by each vehicle. The trajectory prediction module outputs future trajectories for each maneuver class. The vehicle interaction module assigns the true recognized maneuver for each vehicle by combining the confidence values provided by the maneuver recognition module and the feasibility of predicted trajectories given the relative configuration of all vehicles 
}
\label{fig_sys}
\end{figure*}

\noindent \textbf{Data-driven trajectory prediction:}
Data driven trajectory prediction approaches can be broadly classified into clustering based approaches and probabilistic approaches. Clustering based approaches \cite{clust1,clust2,clust3,man1} cluster the training data to give a set of prototype trajectories. Partially observed trajectories are matched with a prototype trajectory based on distance measures such as DTW, LCSS or Hausdorff distance, and the prototype trajectory used as a model for future motion. The main drawback of clustering based approaches is the deterministic nature of the predictions. Probabilistic approaches in contrast, learn a probability distribution over motion patterns and output the conditional distribution over future motion given partial trajectories. These have the added advantage of associating a degree of uncertainty to the future predictions. Gaussian Processes are the most popular approach for modeling trajectories \cite{gp1,gpclass1,gpclass2}. Other approaches include \cite{gmrlat} and \cite{gmrlong} where the authors use Gaussian mixture regression for predicting the longitudinal and lateral motion of vehicles respectively. Of particular interest is the work by Weist \textit{et al.} \cite{vgmm} who use variational Gaussian mixture models (VGMMs) to model the conditional distribution over snippets of trajectory futures given snippets of trajectory history. This approach is much leaner and computationally efficient as compared to Gaussian process regression and was shown to be effective at predicting the highly non-linear motion in turns at intersections. While Weist \textit{et al.} use the velocity and yaw angle of the predicted vehicle obtained from its Differential GPS data, we extend this approach by learning VGMMs for freeway traffic using positions and velocities of surround vehicles estimated using vehicle mounted sensors, similar to our prior work on pedestrian trajectory prediction \cite{self}.\\   

\noindent \textbf{Maneuver-based trajectory prediction: }
Classification of vehicle motion into semantically interpretable maneuver classes has been extensively addressed in both advanced driver assistance systems as well as naturalistic drive studies \cite{man1,man2,man3,man4,jacobmiklas,aida,houenou,gmrlat,gpclass1,gpclass2,schreier,anup}. Most approaches involve using heuristics \cite{houenou} or training classifiers such as SVMs \cite{man3,man4}, HMMs \cite{man1,man2,jacobmiklas,gpclass2}, LSTMs \cite{aida} and Bayesian networks \cite{schreier} using motion based features such as speed, acceleration, yaw rate and other context information such as lane position, turn signals, distance from leading vehicle. 

The works most closely related to our approach are those that use the recognized maneuvers to make predictions of future trajectories. Houenou \textit{et al.} \cite{houenou} classify a vehicle's motion as a keep lane or lane change maneuver based on distance to nearest lane marking and predict the future trajectory by fitting a quintic polynomial between the current motion state of the vehicle and a pre-defined final motion state for each maneuver class. Schreier \textit{et al.} \cite{schreier} classify vehicle motion into one of six different maneuver classes using a Bayesian network based on multiple motion and context based features. A class specific motion model is then defined for each maneuver to generate future trajectories. Most similar in principle to our approach are \cite{gmrlat}, \cite{gpclass1} and \cite{gpclass2} 
where separate probabilistic prediction models are trained for each maneuver class. Tran and Firl \cite{gpclass1} define a separate Gaussian process for three maneuver classes and generate a multi-modal distribution over future trajectories using each model. However, only case based evaluation has been presented. Laugier \textit{et al.} \cite{gpclass2} also define separate Gaussian processes for 4 different maneuvers that are classified using a hierarchical HMM. While they report results for maneuver classification on real highway data, they evaluate trajectory prediction in the context of risk assessment simulated data. Schlechtriemen \textit{et al.} \cite{gmrlat} use a random forest classifier to classify maneuvers into left or right lane changes or keep lane. They use a separate Gaussian mixture regression model for making predictions of lateral movement of vehicles for each class, reporting results on real highway data.
Along similar lines, but without maneuver classes, they also predict longitudinal motion for surround vehicles \cite{gmrlong}. Contrary to this approach, we make predictions for the complete motion of vehicles based on maneuver class, since detection of certain maneuvers like overtakes can help predict both lateral and longitudinal motion of vehicles.\\

\noindent\textbf{Interaction-aware trajectory prediction:}
Relatively few works address the effect of inter-vehicle interaction in trajectory prediction. Kafer \textit{et al.} \cite{vi1} jointly assign maneuver classes for two vehicles approaching an intersection using a polynomial classifier that penalizes cases which would lead to near collisions. Closer to our proposed approach, Lawitzky \textit{et al.} \cite{vi2} consider the much more complex case of assigning maneuver classes to multiple interacting vehicles in a highway setting. However, predicted trajectories and states of vehicle motion are assumed to be given, and results reported using a simulated setting. Contrarily, our evaluation considers the combined complexity due to multiple interacting vehicles as well the difficulty of estimating their future motion. We note that  inter-vehicle interaction is implicitly modeled in \cite{gmrlat} by including relative positions and velocities of nearby vehicles as features for maneuver classification and trajectory prediction.

\section{Overview}

Figure \ref{fig_sys} shows the complete pipeline of our proposed approach. We restrict our setting to purely perception based prediction of surround vehicle motion, without any vehicle-to-vehicle communication. Toward this end, the ego vehicle is equipped with 8 cameras that capture the full surround.  All vehicles within 40 m of the ego vehicle in the longitudinal direction are tracked for motion analysis and prediction. While vehicle tracking is not the focus of this work, we refer readers to a multi-perspective vision based vehicle trackers described in \cite{jacobmiklas,akshay}. The tracked vehicle locations are then projected to the \textit{ground plane} to generate track histories of the surround vehicles in the frame of reference of the ego vehicle. 

The goal of our model is to estimate the future positions and the associated prediction uncertainty for all vehicles in the ego vehicle's frame of reference over the next $t_f$ seconds, given a $t_h$ second snippet of their most recent track histories. The model essentially consists of three interacting modules, namely the \textit{trajectory prediction module}, the \textit{maneuver recognition module} and the \textit{vehicle interaction module}. The trajectory prediction module is the most crucial among the three and can function as a standalone block independent of the remaining two modules. It outputs a linear combination of the trajectories predicted by a motion model that leverages the estimated instantaneous motion of the surround vehicles and a probabilistic trajectory prediction model which learns motion patterns of vehicles on freeways from a freeway trajectory training set. We use constant velocity (CV), constant acceleration (CA) and constant turn rate and velocity (CTRV) models in the interacting multiple model (IMM) framework as the motion models since these capture most instances of freeway motion, especially in light traffic conditions. We use Variational Gaussian Mixture Models (VGMM) for probabilistic trajectory prediction owing to promising results for vehicle trajectory prediction at intersections shown in \cite{vgmm}.

The motion model becomes unreliable for long term trajectory prediction, especially in cases involving a greater degree of decision making by drivers such as overtakes, cut-ins or heavy traffic conditions. These cases are critical from a safety stand-point. However, since these are relatively rare occurrences, they tend to be poorly modeled by a monolithic probabilistic prediction model. Thus we bin surround vehicle motion on freeways into 10 maneuver classes, with each class capturing a distinct pattern of motion that can be useful for future prediction. The intra-maneuver variability of vehicle motion is captured through a VGMM learned for each maneuver class. The maneuver recognition module recognizes the maneuver being performed by a vehicle based on a snippet of it's most recent track history. We use hidden Markov models (HMM) for this purpose. The VGMM corresponding to the most likely maneuver can then be used for predicting the future trajectory. Thus the maneuver recognition and trajectory Prediction modules can be used in conjunction for each vehicle to make more reliable predictions.

Up to this point, our model predicts trajectories of vehicles independent of each other. However the relative configuration of all vehicles in the scene can make certain maneuvers infeasible and certain others more likely, making it a useful cue for trajectory prediction especially in heavy traffic conditions. The vehicle interaction module (VIM) leverages this cue. The maneuver likelihoods and predicted trajectories for the $K$ likeliest maneuvers for each vehicle being tracked are passed to the VIM. The VIM consists of a Markov random field aimed at optimizing an energy function over the discrete space of maneuver classes for all vehicles in the scene. The energy function takes into account the confidence values for all maneuvers given by the HMM and the feasibility of the maneuvers given the relative configuration of all vehicles in the scene. Minimizing the energy function gives the recognized maneuvers and corresponding trajectory predictions for all vehicles in the scene.

\section{Maneuver Recognition Module}
\begin{figure}[t!]
\centering
\includegraphics[width=3.5in]{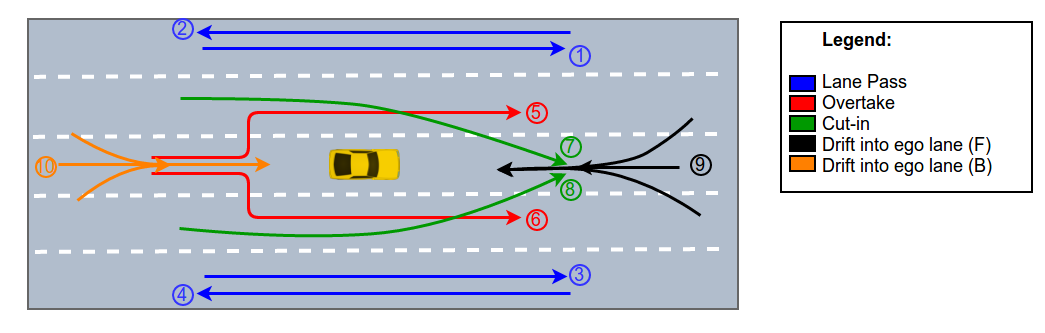}
\caption{\textbf{Maneuver Classes for Freeway Traffic:} We bin the trajectories of surround vehicles in the ego-vehicle frame of reference into 10 maneuver classes: 4 lane pass maneuvers, 2 overtake maneuvers, 2 cut-in maneuvers and 2 maneuvers involving drifting into ego vehicle lane.
}
\label{fig_mans}
\end{figure}
\subsection{Maneuver classes}
\label{manclasses}
We define 10 maneuver classes for surround vehicle motion on freeways in the ego-vehicle's frame of reference. Figure \ref{fig_mans} illustrates the maneuver classes.
\begin{enumerate}
\item \textit{Lane Passes}: Lane pass maneuvers involve vehicles passing the ego vehicle without interacting with the ego vehicle lane. These constitute a majority of the surround vehicle motion on freeways and are relatively easy cases for trajectory prediction owing to approximately constant velocity profiles. We define 4 different lane pass maneuvers as shown in Figure \ref{fig_mans} 
\item \textit{Overtakes}: Overtakes start with the surround vehicle behind the ego vehicle in the ego lane. The surround vehicle changes lane and accelerates in order to pass the ego vehicle. We define 2 different overtake maneuvers, depending on which side the the surround vehicle overtakes.
\item \textit{Cut-ins}: Cut-ins involve a surround vehicle passing the ego vehicle and entering the ego lane in front of the ego-vehicle. Cut-ins and overtakes, though relatively rare, can be critical from a safety stand-point and also prove to be challenging cases for trajectory prediction. We define 2 different cut-ins depending on which side the surround vehicle cuts in from.
\item \textit{Drift into Ego Lane:} Another important maneuver class is when a surround vehicle drifts into the ego vehicle lane in front or behind the ego vehicle. This is also important from a safety standpoint as it directly affects how sharply the ego vehicle can accelerate or decelerate. A separate class is defined for drifts into ego-lane in front and to the rear of the ego vehicle.  
\end{enumerate}

\subsection{Hidden Markov Models}
Hidden Markov models (HMMs) have previously been used for maneuver recognition \cite{man1,man2,jacobmiklas} due to their ability to capture the spatial and temporal variability of trajectories. HMMs can be thought of as combining two stochastic models, an underlying Markov chain of states characterized by state transition probabilities and an emission probability distribution over the feature space for each state. 
The transition probabilities model the temporal variability of trajectories while the emission probabilities model the spatial variability, making HMMs a viable approach for maneuver recognition.

Previous works \cite{man1,jacobmiklas} use HMMs for classifying maneuvers after they have been performed, where the HMM for a particular maneuver is trained using complete trajectories belonging to that maneuver class. In our case, the HMMs need to classify a maneuver based on a small $t_h$ second snippet of the trajectory. Berndt \textit{et al.} \cite{man2} address the problem of maneuver classification based on partially observed trajectories by using only the initial states of a trained HMM to fit the observed trajectory. However, this approach requires prior knowledge of the starting point of the maneuver. In our case, the trajectory snippet could be from any point in the maneuver, and not necessarily the start. We need the HMM to classify a maneuver based on any intermediate snippet of the trajectory. We thus divide the trajectories in our training data into overlapping snippets of $t_h$ seconds and train the maneuver HMMs using these snippets. 

For each maneuver, we train a separate HMM with a left-right topology with only self transitions and transitions to the next state. The state emission probabilities are modeled as mixtures of Gaussians with diagonal covariances. The $x$ and $y$ ground plane co-ordinates and instantaneous velocities in the $x$ and $y$ direction are used as features for training the HMMs. The parameters of the HMMs: the state transition probabilities and the means, variances and weights of the mixture components are estimated using the Baum-Welch algorithm \cite{baum}.

For a car $i$, the HMM for maneuver $k$ outputs the log likelihood:

\begin{equation}
\mathcal{L}_{k}^{i} = \mbox{log}(P(\mathbf{x}^{i}_{h}, \mathbf{y}^{i}_{h}, \mathbf{v_x}^{i}_{h}, \mathbf{v_y}^{i}_{h}|m^{i} = k; \Theta_{k}))
\end{equation}

where $\mathbf{x}^{i}_{h}$, $\mathbf{y}^{i}_{h}$ are the $x$ and $y$ locations of vehicle $i$ over the last $t_h$ seconds and ${ \mathbf{v_x}}^{i}_{h}$, ${\mathbf{v_y}}^{i}_{h}$ are the velocities along the $x$ and $y$ directions over the last $t_h$ seconds. $m^{i}$ is the maneuver assigned to car $i$ and $\Theta_{k}$ are the parameters of the HMM for maneuver $k$

\section{Trajectory Prediction Module} 
The trajectory prediction module predicts the future $x$ and $y$ locations of surround vehicles over a prediction horizon of $t_f$ seconds and assigns an uncertainty to the predicted locations in the form of a $2\times2$ covariance matrix. It averages the predicted future locations and covariances given by both a motion model, and a probabilistic trajectory prediction model. The outputs of the trajectory prediction module for a prediction instant $t_{pred}$ are given by:
\begin{equation}
x_f(t) = \frac{1}{2}\left({x_{f}}_{motion}(t) + {x_{f}}_{prob}(t)\right)
\end{equation}
\begin{equation}
y_f(t) = \frac{1}{2}\left({y_{f}}_{motion}(t) + {y_{f}}_{prob}(t)\right)
\end{equation}
\begin{equation}
\Sigma_f(t) = \frac{1}{2}\left({\Sigma_{f}}_{motion}(t) + {\Sigma_{f}}_{prob}(t)\right)
\end{equation}
 where $t_{pred} \leq t \leq t_{pred}+t_f$

\subsection{Motion Models}
We use the interacting multiple model (IMM) framework for modeling vehicle motion, similar to \cite{imm1}, \cite{imm2}. The IMM framework allows for combining an ensemble of Bayesian filters for motion estimation and prediction by weighing the models with probability values. The probability values are estimated at each time step based on the transition probabilities of an underlying Markov model and how well each model fits the observed motion prior to that time step. We use the following motion models in our ensemble:
\begin{enumerate}
\item \textit{Constant velocity (CV):} The constant velocity models maintains an estimate of the position and velocity of the surround vehicles under the constraint that the vehicles move with a constant velocity. We use a Kalman filter for estimating the state and observations of the CV model. The CV model captures a majority of freeway vehicle motion.
\item \textit{Constant acceleration (CA):} The constant acceleration model maintains estimates of the the vehicle position, velocity and acceleration under the constant acceleration assumption using a Kalman Filter. The CA model can be useful for describing freeway motion especially in dense traffic.
\item \textit{Constant turn rate and velocity (CTRV):} The constant turn rate and velocity model maintains estimates of the the vehicle position, orientation and velocity magnitude under the constant yaw rate and velocity assumption. Since the state update for the CTRV model is non-linear, we use an extended Kalman filter for estimating the state and observations of the CTRV model. The CTRV model can be useful for modeling motion during lane changes  
\end{enumerate}

\subsection{Probabilistic Trajectory Prediction}
We formulate probabilistic trajectory prediction as estimating the conditional distribution:

\begin{equation}
\label{cond}
P(\mathbf{v_{x}}_{f},\mathbf{v_{y}}_{f}|\mathbf{x}_{h},\mathbf{y}_{h},\mathbf{v_{x}}_{h},\mathbf{v_{y}}_{h}, m)
\end{equation}

i.e. the conditional distribution of the vehicle's predicted velocities given the vehicles past positions, velocities and maneuver class. In particular, we are interested in estimating the conditional expected values $[\hat{\mathbf{v_x}}_f;\hat{\mathbf{v_y}}_f]$ and conditional covariance $\Sigma_{v_f}$ of the distribution \ref{cond}. The predicted locations and ${\mathbf{x}_f}_{prob}$, ${\mathbf{y}_f}_{prob}$ can then be obtained by taking the cumulative sum of the predicted velocities, which can be represented using an accumulator matrix $\mathbf{A}$

\begin{equation}
[{\mathbf{x}_f}_{prob};{\mathbf{y}_f}_{prob}] = \mathbf{A}[\hat{\mathbf{v_x}}_f;\hat{\mathbf{v_y}}_f] 
\end{equation}

Similarly, the uncertainty of prediction $\Sigma_{prob}$ can be obtained using the expression:
\begin{equation}
{\Sigma_f}_{prob} = \mathbf{A}\Sigma_{v_f}\mathbf{A}^T 
\end{equation}

\null\quad\\ We use the framework proposed by Weist \textit{et al.} \cite{vgmm} for estimating the conditional distribution \ref{cond}. ($\mathbf{x}_{h}$,$\mathbf{y}_{h}$,$\mathbf{v_{x}}_{h}$,$\mathbf{v_{y}}_{h}$) and ($\mathbf{v_{x}}_{f}$,$\mathbf{v_{y}}_{f}$) are represented in terms of their Chebyshev coefficients, $\mathbf{c}_h$ and $\mathbf{c}_f$. The joint distribution $P(\mathbf{c}_f,\mathbf{c}_h|m)$ for each maneuver class is estimated as the predictive distribution of a variational Gaussian mixture model (VGMM). The conditional distribution $P(\mathbf{c}_f|\mathbf{c}_h,m)$ can then be estimated in terms of the parameters of the predictive distribution. We briefly review the the expressions for $P(\mathbf{c}_f,\mathbf{c}_h|m)$ and $P(\mathbf{c}_f|\mathbf{c}_h,m)$ here. However, the reader is encouraged to refer to \cite{vgmm} for a more detailed treatment.

VGMMs are the Bayesian analogue to standard GMMs, where the model parameters, \{$\mathbf{\pi}$, $\mathbf{\mu}_{1}$, $\mathbf{\mu}_{2}$, ... $\mathbf{\mu}_{K}$, $\mathbf{\Lambda}_{1}$, $\mathbf{\Lambda}_{2}$, ... $\mathbf{\Lambda}_{K}\}$ are given conjugate prior distributions. The prior over mixture weights $\mathbf{\pi}$ is a Dirichlet distribution
\begin{equation}
P(\mathbf{\pi}) = Dir(\mathbf{\pi}|\mathbf{\alpha_{0}})
\end{equation}
The prior over each component mean $\mathbf{\mu}_{k}$ and component precision $\mathbf{\Lambda}_{k}$ is an independent Gauss-Wishart distribution
\begin{multline}
P(\mathbf{\mu}_{k},\mathbf{\Lambda}_{k}) = \mathcal{N}(\mathbf{\mu}_{k}|\mathbf{m}_{0_{k}},(\beta_{0_{k}}\mathbf{\Lambda}_{k})^{-1})\mathcal{W}(\mathbf{\Lambda}_{k}|\mathbf{W}_{0_{k}},\nu_{0_{k}})
\end{multline}
The parameters of the posterior distributions are estimated using the Variational Bayesian Expectation Maximization algorithm \cite{bishop}. The predictive distribution for a VGMM is given by a mixture of Student's t-distributions

\begin{equation}
P(\mathbf{c}_{h},\mathbf{c}_{f}) = \frac{1}{sum(\mathbf{\alpha})}\sum_{k=1}^{K}\alpha_{k}St(\mathbf{c}_{h},\mathbf{c}_{f}|\mathbf{m}_{k},\mathbf{L}_{k},\nu_k+1-d)
\end{equation}
where $d$ is the number of degrees of freedom of the Wishart distribution and 
\begin{equation}
\mathbf{L}_{k} = \frac{(\nu_k+1-d)\beta_{k}}{1+\beta_{k}}\mathbf{W}_{k}
\end{equation}
For a new trajectory history $\mathbf{c_{h}}$, the conditional predictive distribution $P(\mathbf{c_{f}|c_{h}})$ is given by:
\begin{align}
P(\mathbf{c}_{f}|\mathbf{c}_{h}) &= \frac{1}{sum(\mathbf{\hat{\alpha}})}\sum_{k=1}^{K}\hat{\alpha}_{k}St\left(\mathbf{c}_{f}|\mathbf{c}_{h},\mathbf{\hat{m}}_{k},\mathbf{L}_{k},\nu_k+1-d\right)\\ 
\mbox{where}\\
\hat{\nu_{k}} &= \nu_{k} + 1 - d\\
\hat{\alpha_{k}} &= \frac{\alpha_{k}St(\mathbf{c}_{h}|\mathbf{m}_{k,\mathbf{c}_{h}},\mathbf{L}_{k,\mathbf{c}_{h}},\hat{\nu}_{k})}{\sum_{j=1}^{K}\alpha_{j}St(\mathbf{c}_{h}|\mathbf{m}_{j,\mathbf{c}_{h}},\mathbf{L}_{j,\mathbf{c}_{h}},\hat{\nu}_{j})}\\
\hat{\mathbf{m}}_{k} &= \mathbf{m}_{k,\mathbf{c}_{f}} + \mathbf{\Sigma}_{k,\mathbf{c}_{f}\mathbf{c}_{h}}\mathbf{\Sigma}_{k,\mathbf{c}_{h}\mathbf{c}_{h}}^{-1}(\mathbf{c}_{h}-\mathbf{m}_{k,\mathbf{c}_{h}})\\
\hat{\mathbf{L}}_{k}^{-1} &=\frac{\hat{\nu}_{k}}{\hat{\nu}_{k} + d-2}\left(1+\Delta_{k}^{T}\frac{\mathbf{\Sigma}_{k,\mathbf{c}_{h}\mathbf{c}_{h}}}{\hat{\nu}_{k}}\Delta_{k}\right)\mathbf{\Sigma}_{k}^{*}\\
\Delta_{k} &= (\mathbf{c}_{h}-\mathbf{m}_{k,\mathbf{c}_{h}})\\
\mathbf{\Sigma}_{k}^{*} &= \mathbf{\Sigma}_{k,\mathbf{c}_{f}\mathbf{c}_{f}} - \mathbf{\Sigma}_{k,\mathbf{c}_{f}\mathbf{c}_{h}}\mathbf{\Sigma}_{k,\mathbf{c}_{h}\mathbf{c}_{h}^{-1}}\mathbf{\Sigma}_{k,\mathbf{c}_{h}\mathbf{c}_{f}}\\
\mathbf{\Sigma}_{k} &= \frac{\hat{\nu}_{k}+d-2}{\hat{\nu}_{k}+d}\mathbf{L}_{k}^{-1}
\end{align}
 
\section{Vehicle Interaction Module}
The vehicle interaction module is tasked with assigning discrete maneuver labels to all vehicles in the scene at a particular prediction instant based on the confidence of the HMM in each maneuver class and the feasibility of the future trajectories of all vehicles based on those maneuvers given the current configuration of all vehicles in the scene. We set this up as an energy minimization problem. For a given prediction instant, let there be $N$ surround vehicles in the scene with the top $K$ maneuvers given by the HMM being considered for each vehicle. The minimization objective is given by: 

\begin{multline}
\label{en1}
\mathbf{y}^* = \argmin_{\mathbf{y}} \sum_{i=1}^{n}\sum_{k=1}^{K}y_{k}^{i} \left[E^{hmm}_{ik} + \lambda E^{ego}_{ik}\right]\\ +\lambda\sum_{i=1}^{n}\sum_{k=1}^{K}\sum\limits_{\substack{j=1 \\ j\neq i}}^{n}\sum_{l=1}^{K}y_{k}^{i}y_{l}^{j} E^{vi}_{ijkl} 
\end{multline}

s.t.
\begin{align}
\label{en2}
\sum_{k}y_{k}^{i} &= 1 \quad\forall i\\
\label{en3}
y_{k}^{i} &= \begin{cases}
      1, & \text{if car}\ i \text{ is assigned maneuver}\ k \\
      0, & \text{otherwise}
    \end{cases}
\end{align}

The objective consists of three types of energies, the individual Energy terms $E^{hmm}_{ik}$, $E^{ego}_{ik}$ and the pairwise energy terms $E^{vi}_{ijkl}$. The individual energy terms $E^{hmm}_{ik}$ are given by the negative of the log likelihoods provided by the HMM. Higher the confidence of an HMM in a particular maneuver, lower is $-\mathcal{L}_{k}^{i}$ and thus the individual energy term. The individual energy term $E^{ego}_{ik}$ takes into account the interaction between surround vehicles and the ego vehicle. We define the $E^{ego}_{ik}$ as the reciprocal of the closest point of approach for vehicle  $i$ and the ego vehicle over the entire prediction horizon, given that it is performing maneuver $k$, where the ego vehicle position is always fixed to $0$, since it is the origin of the frame of reference. Similarly, the pairwise energy term $E^{vi}_{ijkl}$ is defined as the reciprocal of the minimum distance between the corresponding predicted trajectories for the vehicles $i$ and $j$, assuming them to be performing maneuvers $k$ and $l$ respectively . The terms $E^{ego}_{ik}$ and $E^{vi}_{ijkl}$ penalize predictions where at any point in the prediction horizon, two vehicles are very close to each other. This term leverages the fact that drivers tend to follow paths with low possibility of collisions with other vehicles. The weighting constant $\lambda$ is experimentally determined through cross-validation.  

The minimization objective in the formulation shown in Eq. \ref{en1}, \ref{en2} and \ref{en3} has quadratic terms in $y$ values. In order to leverage integer linear programming for minimizing the energy, we modify the formulation as follows:

\begin{multline}
\mathbf{y}^*,\mathbf{z}^* = \argmin_{\mathbf{y},\mathbf{z}} \sum_{i=1}^{n}\sum_{k=1}^{K}y_{k}^{i}\left[E^{hmm}_{ik} + \lambda E^{ego}_{ik}\right] \\ +\lambda\sum_{i=1}^{n}\sum_{k=1}^{K}\sum\limits_{\substack{j=1 \\ j\neq i}}^{n}\sum_{l=1}^{K}z_{k,l}^{i,j} E^{vi}_{ijkl} 
\end{multline}
s.t.
\begin{align}
\sum_{k}y_{k}^{i} &= 1 \quad\forall i\\
y_{k}^{i} &\in {0,1}\\
z_{k,l}^{i,j} &\leq y_{k}^{i}\\ 
z_{k,l}^{i,j} &\leq y_{l}^{j}\\
z_{k,l}^{i,j} &\geq y_{k}^{i} + y_{l}^{j}-1
\end{align}

This objective can now be optimized using integer linear programming, where the optimal values $\mathbf{y}^*$ give the maneuver assignments for each of the vehicles. These assigned maneuver classes are used by the trajectory prediction module to make future predictions for all vehicles.

\section{Experimental Evaluation}
\subsection{Dataset}

\begin{figure}[t!]
\centering
\includegraphics[width=3.2in]{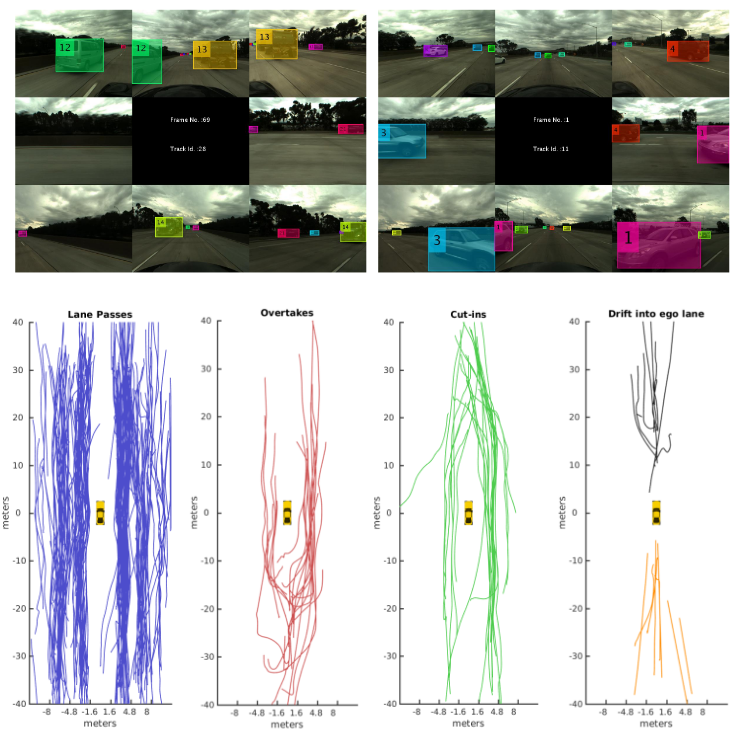}
\caption{\textbf{Dataset:} Examples of annotated frames from the evaluation set (top left and top right) and trajectories belonging to all maneuver classes projected in the ground plane (bottom). We can observe that the trajectory patterns implicitly capture lane information  
}
\label{fig_data}
\end{figure}

\begin{table}[t!]
\centering
\caption{Dataset Statistics}
\label{tab_stats}
\begin{tabular}{@{}lcc@{}}
\toprule
\textbf{Maneuver}                                                      & \textbf{\begin{tabular}[c]{@{}c@{}}Number of \\ trajectories\end{tabular}} & \textbf{\begin{tabular}[c]{@{}c@{}}Number of \\ trajectory snippets\end{tabular}} \\ \midrule
\begin{tabular}[c]{@{}l@{}}Lane Pass \\ (Left Forward)\end{tabular}    & 59                                                                         & 9500                                                                              \\
\begin{tabular}[c]{@{}l@{}}Lane Pass \\ (Left Back)\end{tabular}       & 75                                                                         & 10332                                                                             \\
\begin{tabular}[c]{@{}l@{}}Lane Pass \\ (Right Forward)\end{tabular}   & 110                                                                        & 10123                                                                             \\
\begin{tabular}[c]{@{}l@{}}Lane Pass\\ (Right Back)\end{tabular}       & 48                                                                         & 12523                                                                             \\
Overtake (Left)                                                        & 8                                                                          & 1629                                                                              \\
Overtake (Right)                                                       & 17                                                                         & 2840                                                                              \\
Cut-in (Left)                                                          & 8                                                                          & 1667                                                                              \\
Cut-in (Right)                                                         & 19                                                                         & 3201                                                                              \\
\begin{tabular}[c]{@{}l@{}}Drift into ego \\ lane (Front)\end{tabular} & 11                                                                         & 1317                                                                              \\
\begin{tabular}[c]{@{}l@{}}Drift into ego \\ lane (Rear)\end{tabular}  & 8                                                                          & 553                                                                               \\ \bottomrule
\end{tabular}
\end{table}

We evaluate our framework using real freeway traffic data captured using the testbed described in \cite{testbed}. The vehicle is equipped with 8 RGB video cameras, LIDARs and RADARs synchronously capturing the full surround at a frame rate of 15 fps. Our complete dataset consists of 52 video sequences extracted from multiple drives spanning approximately 45 minutes. The sequences were chosen to capture varying lighting conditions, vehicle types, and traffic density and behavior. 

The 4 longest video sequences, of about 3 minutes each were ground-truthed by human annotators and used for evaluation. Three sequences from the evaluation set represent light to moderate or \textit{free-flowing} traffic conditions, while the remaining sequence represents heavy or \textit{stop-and-go} traffic. The video feed from the evaluation set was annotated with detection boxes and vehicle track-ids for each of the 8 views. All tracks were then projected to the ground plane and assigned a maneuver class label corresponding to the 10 maneuver classes described in Section \ref{manclasses}. If a vehicle track was comprised by multiple maneuvers, the start and end-point of each maneuver was marked. A multi-perspective tracker \cite{jacobmiklas} was used for assigning vehicle tracks for the remaining 48 sequences. These tracks were only used for training the models. Figure \ref{fig_data} shows the track annotations as well as the complete set of trajectories belonging to each maneuver class. Since each trajectory is divided into overlapping snippets of $t_h = 3$ seconds for training and testing our models, we report the data statistics in terms of the total number of trajectories as well as the number of trajectory snippets belonging to each maneuver class in Table \ref{tab_stats}

We report all results using a leave on sequence cross-validation scheme. For each of the 4 evaluation sequences, the HMMs and VGMMs are trained using data from the remaining 3 evaluation sequences as well as the 48 training sequences. Additionally, we use two simple data-augmentation schemes for increasing the size of our training datasets in order to reduce overfitting in the models:
\begin{enumerate}
\item \textit{Lateral inversion:} We flip each trajectory along the lateral direction in the ego frame to give an instance of a different maneuver class. For example, a left cut-in on lateral inversion becomes a right cut in.
\item \textit{Longitudinal shifts:} We shift each of the trajectories by $\pm$ 2, 4 and 6 m in the longitudinal direction in the ego frame to give additional instances of the same maneuver class. We avoid lateral shifts since this would interfere with lane information that is implicitly learned by the probabilistic model.
\end{enumerate}
 
\subsection{Evaluation Measures and Experimental Settings}

\begin{table*}[t!]
\centering
\caption{Quantitative results showing ablative analysis of our proposed model}
\label{tab_results}
\begin{tabular}{@{}cccccccccccc@{}}
\toprule
\multirow{2}{*}{\textbf{Metric}}                                                                    & \textbf{Setting}                                                 & \multicolumn{4}{c}{\textbf{All Trajectories}}                                   & \multicolumn{3}{c}{\textbf{Overtakes and Cut-ins}} & \multicolumn{3}{c}{\textbf{Stop-and-Go Traffic}}                             \\ \cmidrule(lr){2-2}\cmidrule(lr){3-6}\cmidrule(lr){7-9}\cmidrule(lr){10-12} 
                                                                                                    & \begin{tabular}[c]{@{}c@{}}Prediction\\ Horizon (s)\end{tabular} & IMM   & M-VGMM & C-VGMM & \begin{tabular}[c]{@{}c@{}}C-VGMM\\ + VIM\end{tabular} & IMM              & M-VGMM          & C-VGMM         & IMM   & C-VGMM        & \begin{tabular}[c]{@{}c@{}}C-VGMM\\ +VIM\end{tabular} \\ \midrule
\multirow{5}{*}{\textbf{\begin{tabular}[c]{@{}c@{}}Mean\\Absolute\\Error\\(m)\end{tabular}}} & 1                                                                & 0.25 & \textbf{0.24}   & \textbf{0.24}   & \textbf{0.24}                                                     & 0.29            & 0.32            & \textbf{0.29}  & 0.22 & 0.20          & \textbf{0.20}                                         \\ 
                                                                                                    & 2                                                                & 0.72 & 0.70   & \textbf{0.69}   & \textbf{0.69}                                                     & 0.83            & 0.87            & \textbf{0.82}  & 0.68 & 0.65          & \textbf{0.64}                                         \\ 
                                                                                                    & 3                                                                & 1.25 & 1.19   & \textbf{1.18}   & \textbf{1.18}                                                     & 1.47            & 1.46            & \textbf{1.39}  & 1.21 & 1.17          & \textbf{1.14}                                         \\ 
                                                                                                    & 4                                                                & 1.78 & 1.70   & 1.68   & \textbf{1.66}                                                     & 2.17            & 2.05            & \textbf{1.94}  & 1.74 & 1.68          & \textbf{1.65}                                         \\ 
                                                                                                    & 5                                                                & 2.36 & 2.24   & 2.20   & \textbf{2.18}                                                     & 2.90            & 2.68            & \textbf{2.49}  & 2.29 & 2.21          & \textbf{2.17}                                         \\ \midrule

\multirow{5}{*}{\textbf{\begin{tabular}[c]{@{}c@{}}Median\\ Absolute\\ Error\\ (m)\end{tabular}}} & 1                                                                & 0.19 & \textbf{0.17}   & \textbf{0.17}   & \textbf{0.17}                                                     & \textbf{0.23}   & \textbf{0.23}   & \textbf{0.23}  & 0.15 & \textbf{0.13} & \textbf{0.13}                                         \\ 
                                                                                                    & 2                                                                & 0.55 & \textbf{0.52}   & \textbf{0.52}   & \textbf{0.52}                                                     & 0.68            & \textbf{0.65}   & \textbf{0.65}  & 0.48 & 0.46          & \textbf{0.45}                                         \\ 
                                                                                                    & 3                                                                & 0.96 & 0.92   & \textbf{0.91}   & \textbf{0.91}                                                     & 1.24            & 1.13            & \textbf{1.12}  & 0.89 & 0.87          & \textbf{0.83}                                         \\ 
                                                                                                    & 4                                                                & 1.38 & 1.32   & 1.30   & \textbf{1.29}                                                     & 1.92            & 1.71            & \textbf{1.68}  & 1.32 & 1.29          & \textbf{1.27}                                         \\ 
                                                                                                    & 5                                                                & 1.85 & 1.77   & \textbf{1.72}   & \textbf{1.72}                                                     & 2.64            & 2.27            & \textbf{2.12}  & 1.8  & 1.78          & \textbf{1.75}                                                                                      \\ \midrule                                  \multirow{1}{*}{\textbf{\begin{tabular}[c]{@{}c@{}}Class. acc. (\%)\end{tabular}}} &      -                                                            & -  & -     & 83.49   & \textbf{84.24}                                                     & -   & -   & 55.89  & -  & 84.84 & \textbf{87.19} \\ \midrule
                                   \multirow{1}{*}{\textbf{\begin{tabular}[c]{@{}c@{}} Exec. time (s)\end{tabular}}} &  -                                                                & \textbf{0.0346}  & 0.1241    & 0.0891   & 0.1546                                                     & -  & -   & -  & -  & -   & -   \\                                                                                                                                  
                                                                                                     \bottomrule
\end{tabular}
\end{table*}

Our models predict the future trajectory over a prediction horizon of 5 seconds for each 3 second snippet of track history based on the maneuver classified by the HMMs or by the VIM. We use the following evaluation measures for reporting our results:
\begin{enumerate}
\item \textit{Mean Absolute Error}: This measure gives the average absolute deviation of the predicted trajectories from the underlying ground truth trajectories. To compare how the models perform for short term and long term predictions, we report this measure separately for prediction instants up to 5 seconds into the future, sampled with increments of 1 second. The mean absolute error captures the effect of both the number of errors made by the models as well as the severity of the errors.
\item \textit{Median Absolute Error}: We also report the median values of the absolute deviations for up to 5 seconds into the future with 1 second increments, as was done in \cite{gmrlong}. The median absolute error better captures the distribution of the errors made by the models while sifting out the effect of a few drastic errors.   
\item \textit{Maneuver classification accuracy}: We report maneuver classification accuracy for configurations using the maneuver recognition module or the vehicle interaction module.
\item \textit{Execution time}: We report the average execution time per frame, where each frame involves predicting trajectories of all vehicles being tracked at a particular instant. \\
\end{enumerate}

In order to analyze the effect of each of our proposed modules, we compare the trajectory prediction results for following systems

\begin{itemize}
\item \textit{Motion model (\textbf{IMM}):} We use the trajectories predicted by the IMM based motion model as our baseline.\\ 
\item \textit{Monolithic VGMM (\textbf{M-VGMM}):} We consider the trajectories predicted by our trajectory prediction module, where the probabilistic model used is a single monolithic VGMM. This alleviates the need for the maneuver recognition module, since the same model makes predictions irrespective of the maneuver being performed\\
\item \textit{Class VGMMs (\textbf{C-VGMM}):} Here we consider separate VGMMs for each maneuver class in the trajectory prediction module. We use the VGMM corresponding to the maneuver with the highest HMM log likelihood for making the prediction. In this case, maneuver predictions for each vehicle are made independent of the other vehicles in the scene. To keep the comparison with the M-VGMM fair, we use 8 mixture components for each maneuver class for the C-VGMMs, while we use a single VGMM with 80 mixture components for the M-VGMM, ensuring that both models have the same complexity.\\
\item \textit{Class VGMMs with Vehicle Interaction Module (\textbf{C-VGMM + VIM}):} We finally consider the effect of using the vehicle interaction module. In this case, we use the C-VGMMs with the maneuver classes for each of the vehicles in the scene assigned by the vehicle interaction module 
\end{itemize}
We report our results for the complete set of trajectories in the evaluation set. Additionally, we also report results on the subsets of overtake and cut-in maneuvers and stop-and-go traffic. Since overtakes and cut-ins are rare safety critical maneuvers with significant deviation from uniform motion, these are challenging cases for trajectory prediction. Similarly, due to the high traffic density in stop-and-go scenarios, vehicles affect each others motion to a much greater extent as compared to free-flowing traffic, making it a challenging scenario for trajectory prediction. 

\begin{figure*}
\centering
\begin{subfigure}[b]{.3\linewidth}
\includegraphics[width=\linewidth]{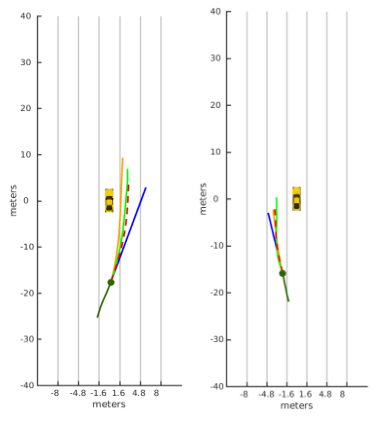}
\caption{}\label{fig_ovNL}
\end{subfigure}
\begin{subfigure}[b]{.3\linewidth}
\includegraphics[width=\linewidth]{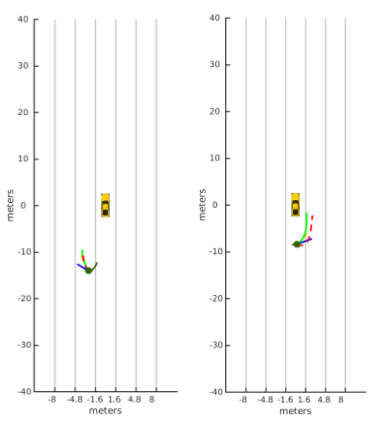}
\caption{}\label{fig_ovE}
\end{subfigure}

\begin{subfigure}[b]{.3\linewidth}
\includegraphics[width=\linewidth]{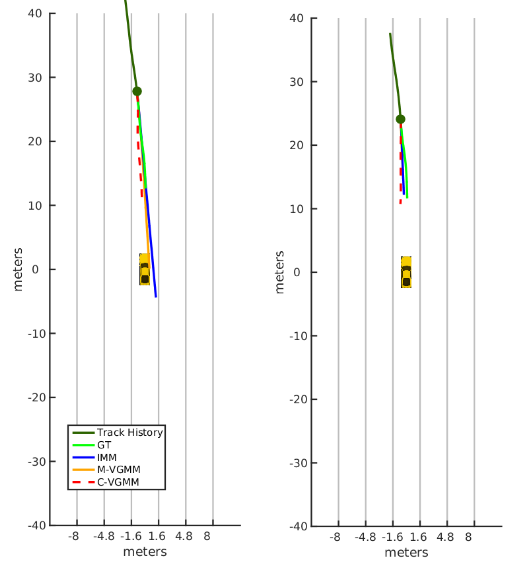}
\caption{}\label{fig_decel}
\end{subfigure}
\begin{subfigure}[b]{.3\linewidth}
\includegraphics[width=\linewidth]{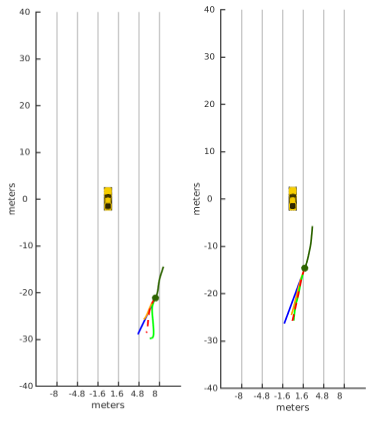}
\caption{}\label{fig_laneI}
\end{subfigure}
\caption{\textbf{Analysis of predictions made by CV, M-VGMM and C-VGMM models:} (a): Better prediction of lateral motion in overtakes by the probabilistic models. (b): Early detection of overtakes by the HMM. (c): Deceleration near the ego vehicle predicted by the C-VGMM. (d): Effect of lane information implicitly encoded by the M-VGMM and C-VGMM}
\label{fig_cases}
\end{figure*}

\begin{figure*}
\centering
\begin{subfigure}[b]{0.7\linewidth}
\includegraphics[width=\linewidth]{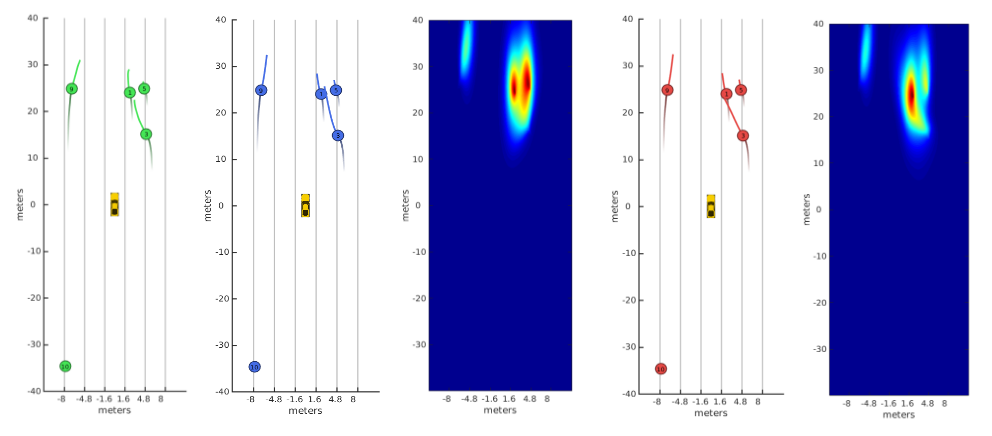}
\caption{Infeasible lane pass is correctly changed to cut-in}\label{fig_vi1}
\end{subfigure}

\begin{subfigure}[b]{0.7\linewidth}
\includegraphics[width=\linewidth]{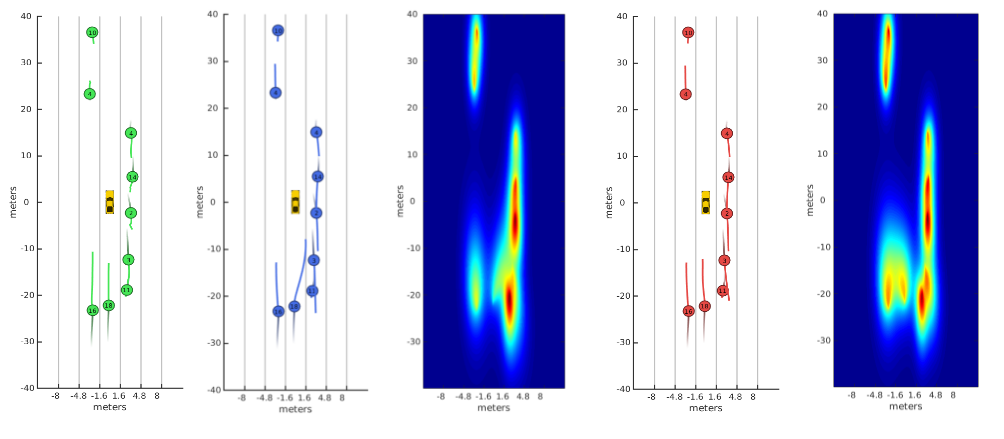}
\caption{Infeasible overtake correctly changed to tail-gating}\label{fig_vi2}
\end{subfigure}

\begin{subfigure}[b]{0.7\linewidth}
\includegraphics[width=\linewidth]{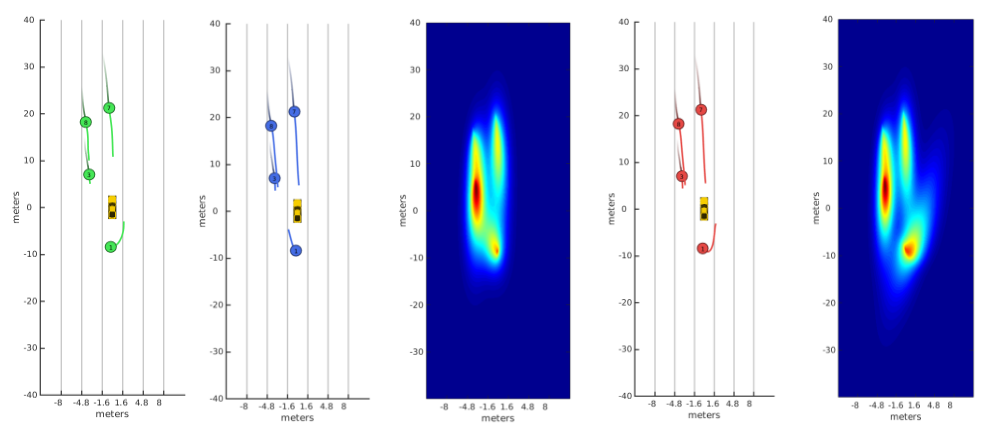}
\caption{Infeasible left overtake changed to the correct overtake direction}\label{fig_vi3}
\end{subfigure}

\caption{\textbf{Case Studies analyzing the effect of the VIM:} Each case shows from left to right: The ground truth, predictions made independently for each vehicle, uncertainty of the independent predictions, predictions made with the VIM, uncertainties of the VIM predictions}
\label{fig_vi}
\end{figure*}

\subsection{Ablative Analysis}
Table \ref{tab_results} shows the quantitative results of our ablation experiments. We note from the results on the complete evaluation set that the probabilistic trajectory prediction models outperform the baseline of the IMM. The M-VGMM has lower values for both mean as well as median absolute error as compared to the IMM suggesting that the probabilistic model makes fewer as well as less drastic errors on an average. We get further improvements in mean and median absolute deviations using the C-VGMMs suggesting that subcategorizing trajectories into maneuver classes leads to a better probabilistic prediction model.

This is further highlighted based on the prediction results for the challenging maneuver classes of overtakes and cut-ins. We note that the C-VGMM significantly outperforms the CV and M-VGMM models both in terms of mean and median absolute deviation for overtakes and cut-ins. This trend becomes more pronounced as the prediction horizon is increased. This suggests that the motion model is more error prone due to the non-uniform motion in overtakes and cut-ins while these rare classes get underrepresented in the distribution learned by the monolithic M-VGMM. Both of these issues get addressed through the C-VGMM.  We analyze this further by considering specific cases of predictions made by the IMM, M-VGMM and C-VGMM in Section \ref{anal1}

Comparing the maneuver classification accuracies for the case of C-VGMM and C-VGMM + VIM, we note that the VIM corrects some of the maneuvers assigned by the HMM. This in turn leads to improved trajectory prediction as seen from the mean and median absolute error values. We note that this effect is more pronounced in case of stop-and-go traffic, since the dense traffic conditions cause more vehicles to affect each others motion leading to a greater proportion of maneuver class labels to be re-assigned by the VIM. Section \ref{anal2} analyses cases where the VIM reassigns maneuver labels assigned by the HMM due to the relative configuration of all vehicles in the scene.

\subsection{Analysis of execution time}
Table \ref{tab_results} also shows the average execution time per frame for the 4 system configurations considered. As expected, the IMM baseline has the lowest execution time since all other configurations build upon it. We note that the C-VGMM runs faster than the M-VGMM in spite of having the overhead of the HMM based maneuver recognition module. This is because the M-VGMM is a much bulkier model as compared to any single maneuver C-VGMM. Thus in spite of involving an extra step, the maneuver recognition module allows us to choose a much leaner model, effectively reducing the execution time while improving performance. The VIM is a more time intensive overhead and almost doubles the run time of the C-VGMM. However, even in it's most complex setting, the proposed framework can be deployed at a frame rate of almost 6 fps, which is more than sufficient for the application being considered.

\subsection{Analyzing predictions of IMM, M-VGMM and C-VGMM models}
\label{anal1}
Figure \ref{fig_cases} shows the trajectories predicted by the CV, M-VGMM and C-VGMM models for 8 different instances.

Figure \ref{fig_ovNL} shows two prediction instants where the vehicle is just about to start the non-linear part of overtake maneuvers. We observe that the IMM makes erroneous predictions in both cases. However, both the M-VGMM and C-VGMM manage to predict the non-linear future trajectory. 

Figure \ref{fig_ovE} shows two prediction instants in the early part of overtake maneuvers. We note that both the IMM and M-VGMM make errors in prediction. However the position of the surround vehicle along with the slight lateral motion provide enough context to the maneuver recognition module to detect the overtake maneuver early. Thus, the C-VGMM manages to predict that the surround vehicle would move to the adjacent lane and accelerate in the longitudinal direction, although there is no such cue from the vehicles existing state of motion  

Figure \ref{fig_decel} shows two instants the trajectory of a vehicle that decelerates as it approaches the ego vehicle from the front. This trajectory corresponds to the drift into ego-lane maneuver class. In the first case (left), the vehicle has not started decelerating, causing the IMM to assign a high probability to the CV model. The IMM thus predicts the vehicle to keep moving at a constant velocity and come dangerously close to the ego vehicle. Similarly, the M-VGMM makes a poor prediction since these maneuvers are underrepresented in the training data. The C-VGMM however manages to correctly predict the surround vehicle to decelerate. In the second case (right), we observe that the car has already started decelerating. This allows the IMM to assign a greater weight to the CA model and correct its prediction

Finally Figure \ref{fig_laneI} shows two interesting  instances of the lane pass right back maneuver that is well represented in the training data. The vehicle makes a lane change in both of these instances. We note, as expected, that the IMM poorly predicts these trajectories. However both the M-VGMM and C-VGMM correctly predict the vehicle to merge into the lane, suggesting that the probabilistic models may have implicitly encoded lane information.

\subsection{Vehicle Interaction Model Case Studies}
\label{anal2}

Figure \ref{fig_vi} shows three cases where the recognized maneuvers and predicted trajectories are affected by the VIM. In each case, the green plots show the ground truth of future tracks, the blue plots show the predictions made for each vehicle independently and the red plots show the predictions based on the VIM. Additionally we plot the prediction uncertainties for either case.  

Consider  the first case in Figure \ref{fig_vi1}, in particular vehicle 3. We note from the blue plot that the HMM predicts the vehicle to perform a lane pass. However the the vehicle's path forward is blocked by vehicles 1 and 5. The VIM thus infers vehicle 3 to perform a cut-in in with respect to the ego-vehicle in order to overtake vehicle 5.

In Figure \ref{fig_vi2}, the HMM predicts vehicle 18 to overtake the ego-vehicle from the right. However, we can see that the right lane is occupied by vehicles 11, 3 and 2. These vehicles yield high values of pairwise energies with vehicle 18 for the overtake maneuver. The VIM thus correctly manages to predict that vehicle 18 would end up tail-gating by assigning it the maneuver drift into ego lane (rear). 

Finally Figure \ref{fig_vi3} shows a very interesting case where the HMM predicts vehicle 1 to overtake the ego vehicle from the left. Again, the left lane is occupied by other vehicles making the overtake impossible to execute from the left. However, compared to the previous case, these vehicles are slightly further away and can be expected to yield relatively smaller energy terms as compared to case (b). However, these terms are enough to offset the very slight difference in the HMM's confidence values between the left and right overtake since both maneuvers do seem plausible if we consider vehicle 1 independently. Thus the VIM reassigns the maneuver for vehicle 1 to a right overtake, making the prediction closely match the ground truth.

\section{Conclusions}
In this paper, we have presented a unified framework for surround vehicle maneuver recognition and motion prediction using vehicle mounted perceptual sensors, that leverages the instantaneous motion of vehicles, an understanding of motion patterns of freeway traffic and the effect of inter-vehicle interactions. The proposed framework outperforms an interacting multiple model based trajectory prediction baseline and runs in real time at about 6 frames per second.  

An ablative analysis for the relative importance of each cue for trajectory prediction has been presented. In particular, we have shown that probabilistic modeling of surround vehicle trajectories is a more versatile approach, and leads to better predictions as compared to a purely motion model based approach for many safety critical trajectories around the ego vehicle. Additionally, subcategorizing trajectories based on maneuver classes leads to better modeling of motion patterns. Finally, incorporating a model that takes into account interactions between surround vehicles for simultaneously predicting each of their motion leads to better prediction as compared to predicting each vehicle's motion independently.

The proposed approach could be treated as a general framework, where improvements could be made to each of the three interacting modules.


\section*{Acknowledgments}
We would like to thank our colleague Kevan Yuen for his invaluable contribution in the design and development of the testbed used in this work, and its software system. We are pleased to acknowledge the support of our research by various sponsoring agencies. Finally, we would like to thank the anonymous reviewers for their valuable feedback.

\ifCLASSOPTIONcaptionsoff
  \newpage
\fi

\newpage
\begin{IEEEbiography}[{\includegraphics[width=1in,height=1in,clip,keepaspectratio]{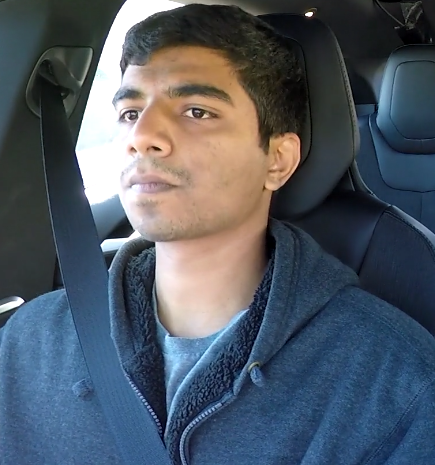}}]{Nachiket Deo}
is currently working towards his PhD in electrical engineering from the University of California at San Diego (UCSD), with a focus on intelligent systems, robotics, and control. His research interests span computer vision and machine learning, with a focus on motion prediction for vehicles and pedestrians
\end{IEEEbiography}
 \vspace{-80 mm} 
\begin{IEEEbiography}[{\includegraphics[width=1in,height=1.25in,clip,keepaspectratio]{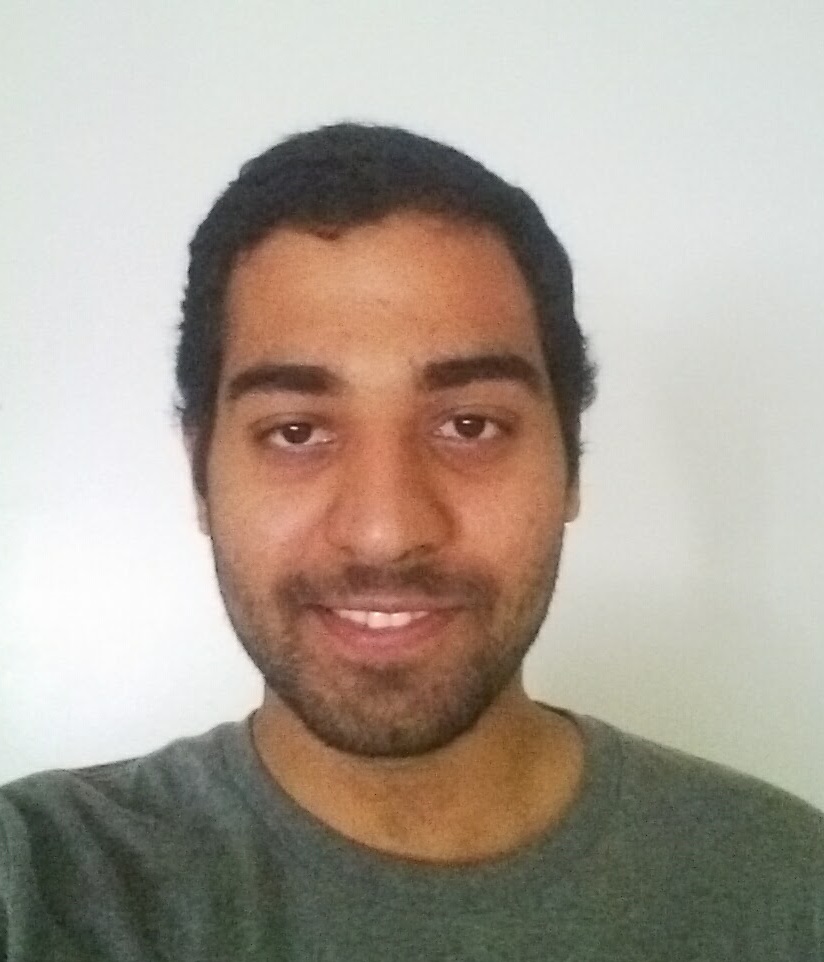}}]{Akshay Rangesh}
is currently working towards his PhD in electrical engineering from the University of California at San Diego (UCSD), with a focus on intelligent systems, robotics, and control. His research interests span computer vision and machine learning, with a focus on object detection and tracking, human activity recognition, and driver safety systems in general. He is also particularly interested in sensor fusion and multi-modal approaches for real time algorithms.
\end{IEEEbiography}
 \vspace{-80 mm} 
\begin{IEEEbiography}[{\includegraphics[width=1in,height=1.25in,clip,keepaspectratio]{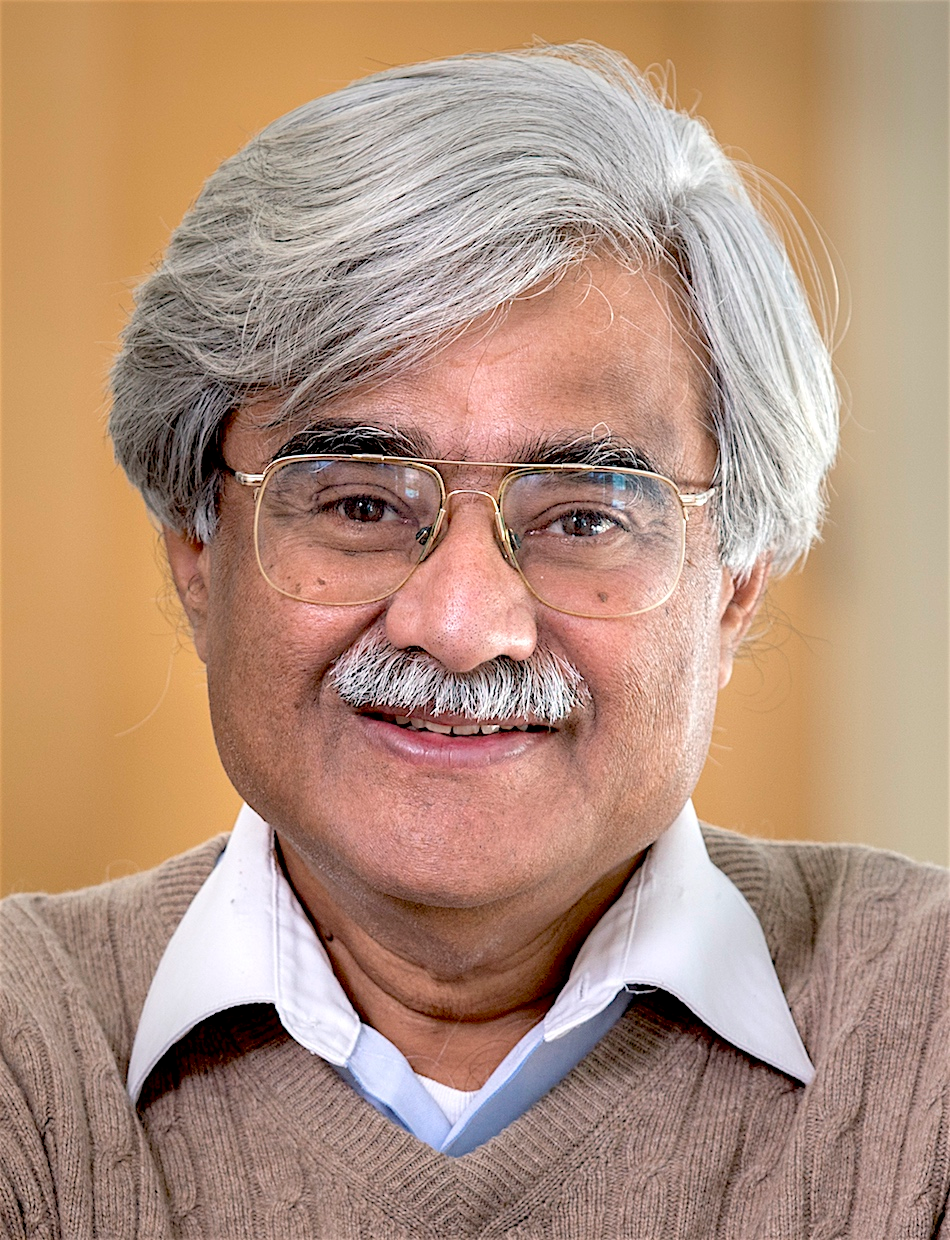}}]{Mohan Manubhai Trivedi}
is a Distinguished Professor at University of California, San Diego (UCSD) and the founding director of the UCSD LISA: Laboratory for Intelligent and Safe Automobiles,
winner of the IEEE ITSS Lead Institution Award (2015). Currently, Trivedi and his team
are pursuing research in intelligent vehicles, autonomous driving, machine perception, machine learning, human-robot interactivity, driver assistance. Three of his students have received "best dissertation" recognitions and over twenty best papers/finalist recognitions. Trivedi is a Fellow of IEEE, ICPR and SPIE. He received the IEEE ITS Society's highest accolade "Outstanding Research Award" in 2013. Trivedi serves frequently as a consultant to industry and government agencies in the USA and abroad. 
\end{IEEEbiography}

\end{document}